\documentclass[]{article}
\usepackage[a4paper]{geometry}
\usepackage{mtsummit2023}
\usepackage{times}
\usepackage{url}
\usepackage{latexsym}
\usepackage{natbib}
\usepackage{layout}
\usepackage[hang,marginal]{footmisc}

\usepackage{verbatim}
\usepackage{booktabs}
\usepackage{multicol}
\usepackage{multirow}
\usepackage{xcolor}
\usepackage{graphicx}
\usepackage{amsmath}
\usepackage{subfigure}

\begin{document}

\title{\bf A Case Study on Context Encoding in Multi-Encoder based Document-Level Neural Machine Translation}

\author{\name{\bf Ramakrishna Appicharla} \hfill  \addr{appicharla\_2021cs01@iitp.ac.in}\\
        \name{\bf Baban Gain} \hfill \addr{gainbaban@gmail.com }\\
        \addr{Department of Computer Science and Engineering, \\Indian Institute of Technology Patna,
        Patna, India}
\AND
       \name{\bf Santanu Pal} \hfill \addr{santanu.pal.ju@gmail.com }\\
        \addr{Wipro AI, Lab45, London, UK}
\AND
       \name{\bf Asif Ekbal} \hfill \addr{asif@iitp.ac.in}\\
        \addr{Department of Computer Science and Engineering, \\Indian Institute of Technology Patna,
        Patna, India}
}

\maketitle
\pagestyle{empty}

\begin{abstract}
Recent studies have shown that the multi-encoder models are agnostic to the choice of context, and the context encoder generates noise which helps improve the models in terms of BLEU score. In this paper, we further explore this idea by evaluating with context-aware pronoun translation test set by training multi-encoder models trained on three different context settings \textit{viz,} previous two sentences, random two sentences, and a mix of both as context. Specifically, we evaluate the models on the ContraPro test set to study how different contexts affect pronoun translation accuracy. The results show that the model can perform well on the ContraPro test set even when the context is random. We also analyze the source representations to study whether the context encoder generates noise. Our analysis shows that the context encoder provides sufficient information to learn discourse-level information. Additionally, we observe that mixing the selected context (the previous two sentences in this case) and the random context is generally better than the other settings.
\end{abstract}

\section{Introduction}

Document-level neural machine translation (DocNMT) has gained a lot of attention due to the ability to incorporate context through different paradigms such as single encoder \citep{tiedemann-scherrer-2017-neural, agrawal2018contextual}, multiple encoders \citep{zhang-etal-2018-improving, li-etal-2020-multi-encoder, huo-etal-2020-diving}, memory networks \citep{maruf-haffari-2018-document} and pre-trained language models \citep{donato-etal-2021-diverse}. This additional context helps to produce more consistent translations \citep{bawden-etal-2018-evaluating, voita-etal-2019-good} than sentence-level models. Two of the most followed approaches to incorporating context are the concatenation-based and multi-encoder-based approaches. In the concatenation-based method, by concatenating context and current input sentence, a context-aware input sentence is generated \citep{tiedemann-scherrer-2017-neural, agrawal2018contextual, junczys-dowmunt-2019-microsoft, zhang-etal-2020-long} and use it as the input to the encoder. In the multi-encoder approach, to encode the source or target context, an additional encoder is used \citep{zhang-etal-2018-improving, voita-etal-2018-context, kim-etal-2019-document, ma-etal-2020-simple} and the entire model is jointly optimized. Typically, the current sentence's neighboring sentences (previous or next) are used as the context, whereas models consist of multiple encoders and a single decoder.

Recent studies on Multi-Encoder (MultiEnc) based DocNMT models \citep{li-etal-2020-multi-encoder, wang-etal-2020-tencent, gain-etal-2022-investigating} have shown that the context-encoder is acting as noise generator which improves the robustness of the model and makes the model agnostic to the choice of context. However, the improvement is in terms of BLEU \citep{papineni-etal-2002-bleu}, which might not capture the discourse-level phenomenon effectively \citep{muller-etal-2018-large}. This phenomenon is not studied well in the existing literature. The context encoder might not generate noise if the model can effectively capture any discourse phenomenon, such as pronoun translation from the source to the target language. Modeling the relation between sentences in a given document is essential to capture any discourse phenomenon \citep{voita-etal-2018-context}. To this end, we hypothesize that, during the training phase, if the model can learn the similarities between all the sentences in a given document given in the form of (\textit{context}, \textit{source}) pairs, the context encoder might not be generating noise since all the sentences in the given document are connected via the context.

In this work, we aim to study the effect of the context in MultiEnc-based DocNMT models and the models' behavior in random context settings but not to introduce a novel technique. We use the `Outside Attention Multi-Encoder' model \citep{li-etal-2020-multi-encoder} with four different context settings to study the effect of the context. We conduct experiments on News-commentary v14 and TED corpora from English--German direction. We report the results on the ContraPro test set \citep{muller-etal-2018-large}, a contrastive test set to evaluate models' performance in translating pronouns. We also report sentence-BLEU (s-BLEU) \citep{papineni-etal-2002-bleu}, document-BLEU (d-BLEU) \citep{liu-etal-2020-multilingual-denoising, bao-etal-2021-g}, and COMET \citep{rei-etal-2020-comet} scores.

To summarize, the specific attributes of our current work are as follows:

\begin{itemize}
    \item We conduct experiments on multi-encoder based DocNMT models to study if the context encoder is generating noise or not by evaluating the model with ContraPro \citep{muller-etal-2018-large} test set.
    \item We empirically show that the model can learn discourse-level information even when trained with random context.
\end{itemize}

\section{Related Work}

The performance of document-level NMT is better than that of sentence-level NMT models due to the encoding of context \citep{sim-smith-2017-integrating, voita-etal-2018-context}. Towards this goal to represent context, \citet{tiedemann-scherrer-2017-neural} concatenate consecutive sentences and use them as input to the single-encoder-based DocNMT model. \citet{agrawal2018contextual} conducted experiments on varying neighboring contexts and then tied them with the current sentence as input to their model. However, this approach introduced a lot of long-range dependencies. This problem can be alleviated by introducing an additional encoder to encode the context. Towards this, \citet{zhang-etal-2018-improving} and \citet{voita-etal-2018-context} proposed transformer-based multi-encoder NMT models where the other encoder is used to encode the context. While \citet{miculicich-etal-2018-document} proposed a hierarchical attention network to encode the context and a more recent approach \citet{kang-etal-2020-dynamic} proposed a reinforcement learning-based dynamic context selection module for DocNMT. Recent studies \citep{kim-etal-2019-document, li-etal-2020-multi-encoder, wang-etal-2020-tencent} have shown that the improvement in the performance of multi-encoder DocNMT models is not due to context encoding but rather the context encoder acting as a noise generator, which improves the robustness of the DocNMT model. Other approaches such as pre-training \citep{junczys-dowmunt-2019-microsoft, donato-etal-2021-diverse} and memory-based approaches \citep{feng-etal-2022-learn} are shown to have improved the performance of DocNMT models. Recent studies \citep{sun-etal-2022-rethinking, post2023escaping} have demonstrated that single encoder-decoder models can capture long-range dependencies. Still, special care must be taken to break a large document into smaller fragments for training. Along with document-level translation, multi-encoder models are also commonly used in automatic post-editing \citep{pal-etal-2019-usaar, pal-etal-2018-transformer, junczys-dowmunt-grundkiewicz-2018-ms, shin-lee-2018-multi}, multimodal translation \citep{libovicky-etal-2018-input, liu-etal-2020-meet} and multitask learning \citep{luong2015multi, zhang2017me, anastasopoulos-chiang-2018-tied} scenarios.

In this work, we study the effect of context in the multi-encoder \citep{li-etal-2020-multi-encoder} based approach. Specifically, we verify if the context encoder is generating noise or not. If the context encoder is generating noise, then the model may not be able to effectively capture the discourse-level phenomenon, such as pronoun translation accuracy, even when the model can perform well on other automatic metrics, such as BLEU.

\section{Methodology}

\subsection{Outside Context Multi-Encoder Model}

\begin{figure}[!ht]
    \centering
    \includegraphics[scale=0.7]{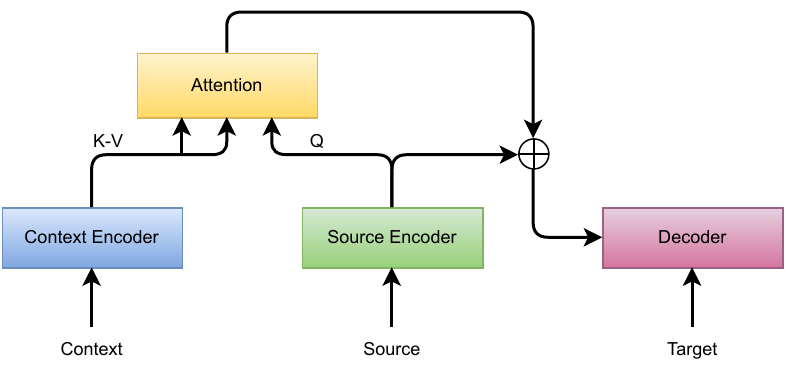}
    \caption{The overview of the Outside Context Multi-Encoder DocNMT architecture. The input to the model consists of (\textit{Context}, \textit{Source}, \textit{Target}, \textit{Label}). Both the encoders are encoding \textit{Context} and \textit{Source}. The Context and Source encoder outputs are passed through the Attention layer. Here, `\textit{K-V}' represents Key-Value pairs from the Context encoder, and `\textit{Q}' represents Query from the Source encoder. The output of the Attention layer is element-wise summed with the output of the Source encoder before passing to the Decoder. None of the layers are shared.}
    \label{fig: outside-attn}
\end{figure}

We conduct all experiments on the `Outside Attention Multi-Encoder' \citep{li-etal-2020-multi-encoder} model. The model (cf. Fig \ref{fig: outside-attn}) consists of two encoders and one decoder. Both source and context are encoded through two encoders, and the output of these encoders is passed through an attention layer. An element-wise addition is performed on the outputs of the source encoder and the attention layer before passing it to the decoder.

\subsection{Context-Aware Models}

We train context-aware models in four different settings. They are,

\begin{enumerate}
    \item \textbf{MultiEnc-Prev@2}: In this setting, the context consists of the previous two sentences concatenated, with respect to the current source sentence \citep{zhang-etal-2018-improving}, and the model is trained in this context setting. The same sentence is used as context if the sentence is the first or second sentence in the document.
    \item \textbf{MultiEnc-Random@2}: In this setting, the context consists of two random sentences sampled from the complete training set.
    \item \textbf{MultiEnc-Mix@2}: In this setting, 50\% of the training set consists of context from `\textit{MultiEnc-Prev@2}' setting and the remaining 50\% consists of context from `\textit{MultiEnc-Random@2}' setting. Essentially, this setting combines the context settings from the above two approaches.
    \item \textbf{MultiEnc-Mix-Adapt@2}: This setting is similar to the `\textit{MultiEnc-Mix@2}' setting but the loss during the training is modified as follows:

    \begin{equation}
        \mathcal{L}\, =\, \alpha\, \times\, \mathcal{L}
    \end{equation}
    Where `$\alpha$' is a scaling factor which is the fraction of source sentences having the previous two sentences as context over all the sentences in the current batch, and `$\mathcal{L}$' is the loss for the current batch. During the training, we also provide the labels list to facilitate this counting, with 0 indicating random context and 1 indicating the previous two sentences as context. Our motivation in this approach is to penalize the model \footnote{If the entire batch consists of random context, then the loss for that batch will be 0 and vice versa.} based on the number of random context inputs per batch and force the model not to learn from random context.
\end{enumerate}

The validation set consists of the previous two sentences as the context in all four settings.

\section{Experimental Setup}

This section describes the data sets and the experimental setup used in the experiments.

\subsection{Data Statistics}

We conduct experiments on English--German corpus obtained from combining \footnote{We combine corpora from all three sources into a single corpus and train our models on this corpus.} WMT news-commentary, IWSLT`17 TED, and Europarl-v7 corpora. For the WMT news-commentary, we use news-commentary v14\footnote{\url{https://data.statmt.org/news-commentary/v14/training/}} as the train set and newstest2018 as the test set. For IWSLT`17 TED and Europarl-v7 corpora, we follow the train and test set splits mentioned in the previous work \citep{maruf-etal-2019-selective}\footnote{\url{https://github.com/sameenmaruf/selective-attn/tree/master/data}}. We use newstest2017 as the validation set for all the models. The models are trained from English to German. Table~\ref{tab: corpora_stats} shows data statistics of the train, validation, and test sets.

\begin{table}[!ht]
    \centering
    \small
    \begin{tabular}{llcc}
        \toprule
        \textbf{Data} & \textbf{Corpus} & \textbf{\# Sentences} & \textbf{\# Documents} \\
        \cmidrule{1-4}
        \multirow{4}*{\textbf{Train}} & News & 329,041 & 8,462 \\
                       & TED & 206,112 & 1,698 \\
                       & Europarl & 1,666,904 & 117,855 \\
                       \cmidrule{2-4}
                       & \textbf{Total} & 2,202,057 & 128,015 \\
        \midrule
        \textbf{Validation} & newstest2017 & 3,004 & 130 \\
        \midrule
        \multirow{3}*{\textbf{Test}} & News & 2,998 & 122 \\
                                     & TED & 2,271 & 23 \\
                                     & Europarl & 5,134 & 360 \\
        \bottomrule
    \end{tabular}
    \caption{Data statistics of corpora. \textbf{\# Sentences}, \textbf{\# Documents} represent the number of sentences and documents, respectively. The train set consists of the corpus obtained by combining News, TED, and Europarl corpora. The models are tested on each test set separately.}
    \label{tab: corpora_stats}
\end{table}

\subsection{NMT Model Setups}

We conduct all the experiments on transformer architecture \citep{vaswani2017attention}. All the models are implemented in PyTorch\footnote{\url{https://pytorch.org/}}. The models consist of 6-layer encoder-decoder stacks, 8 attention heads, and a 2048-cell feed-forward layer. Positional and token embedding sizes are set to 512. Adam optimizer \citep{DBLP:journals/corr/KingmaB14} is used for training with a noam learning rate scheduler \citep{vaswani2017attention} and the initial learning rate set to 0.2. The dropout and warmup steps are set to 0.3 \citep{li-etal-2020-multi-encoder, sun-etal-2022-rethinking} and 16,000 \citep{popel2018training} respectively, and we used a mini-batch of 30 sentences. We create joint subword vocabularies of size 40,000 by combing source and target parts of the training corpus into a single joint corpus. We use the BPE \citep{sennrich-etal-2016-neural} to create subword vocabularies with SentencePiece \citep{kudo-richardson-2018-sentencepiece} implementation. We also learn the positional encoding of tokens \citep{devlin-etal-2019-bert}, and the maximum sequence length is set to 140 tokens for all models. All models are trained till convergence, and we use perplexity on the validation set as early stopping criteria with the patience of 7 \citep{popel2018training}. We report results on the best model checkpoint saved during the training. We perform beam search during inference with beam size 4 and length penalty of 0.6 \citep{45610}.

\section{Results and Analysis}

We test context-aware models with two different contexts \textit{viz.} previous two and random sentences as context. Table \ref{tab: main_results} shows the s-BLEU \citep{papineni-etal-2002-bleu, post-2018-call}\footnote{sacreBLEU signature: $\text{nrefs:1}|\text{case:mixed}|\text{eff:no}|\text{tok:13a}|\text{smooth:exp}|\text{version:2.3.1}$}, d-BLEU \citep{liu-etal-2020-multilingual-denoising, bao-etal-2021-g}, and COMET \citep{rei-etal-2020-comet}\footnote{COMET model: wmt22-comet-da} scores. Overall, \textit{MultiEnc-Mix@2} and \textit{MultiEnc-Mix-Adapt@2} models achieve the best overall scores on all test sets. The \textit{MultiEnc-Mix@2} model achieves 23.1 s-BLEU and 25.3 d-BLEU on the News test set in \textit{Prev@2} setting. For Ted test set, \textit{MultiEnc-Mix-Adapt@2} model achieving 20.8 s-BLEU and 24.6 d-BLEU in \textit{Prev@2} setting. For the Europarl test set, \textit{MultiEnc-Mix@2} and \textit{MultiEnc-Mix-Adapt@2} models are achieving 26.5 s-BLEU and the \textit{MultiEnc-Prev@2} model achieving 28.8 d-BLEU on \textit{Random@2} setting. However, the results from \textit{Prev@2} and \textit{Random@2} settings are very similar and not statistically significant when compared to each other.

\begin{table}[!ht]
    \centering
    \resizebox{1.0\linewidth}{!}{
    \begin{tabular}{lccccccccccc}
        \toprule
        \multirow{2}*{\textbf{Model}} & \multicolumn{3}{c}{\textbf{s-BLEU}} & & \multicolumn{3}{c}{\textbf{d-BLEU}} & & \multicolumn{3}{c}{\textbf{COMET}} \\
        \cmidrule{2-4}\cmidrule{6-8}\cmidrule{10-12}
         & News & TED & Europarl & & News & TED & Europarl & & News & TED & Europarl \\
        \midrule
        \multicolumn{12}{c}{\textbf{Prev@2}} \\
        \midrule
        MultiEnc-Prev@2 & 22.9 & 20.4 & 26.3 & & 25.2 & 24.3 & 28.6 & & 65.3 & 71.3 & 81.9 \\
        MultiEnc-Random@2 & 22.7 & 19.9 & 26.4 & & 25.0 & 23.8 & 28.7 & & 65.4 & 70.7 & 81.9 \\
        MultiEnc-Mix@2 & \textbf{23.1} & 20.3 & 26.4 & & \textbf{25.3} & 24.2 & 28.7 & & \textbf{65.8} & 71.1 & \textbf{82.1} \\
        MultiEnc-Mix-Adapt@2 & 22.9 & \textbf{20.8} & 26.4 & & 25.1 & \textbf{24.6} & 28.7 & & 65.5 & 71.3 & 82.0 \\
        \midrule
        \multicolumn{12}{c}{\textbf{Random@2}} \\
        \midrule
        MultiEnc-Prev@2 & 22.9 & 20.5 & 26.4 & & 25.2 & 24.4 & \textbf{28.8} & & 65.1 & 71.3 & 81.8 \\
        MultiEnc-Random@2 & 22.7 & 19.9 & 26.4 & & 25.0 & 23.8 & 28.7 & & 65.4 & 70.7 & 81.9 \\
        MultiEnc-Mix@2 & 23.0 & 20.5 & \textbf{26.5} & & 25.2 & 24.4 & 28.7 & & 65.6 & 71.1 & 82.0 \\
        MultiEnc-Mix-Adapt@2 & 22.7 & 20.6 & \textbf{26.5} & & 25.0 & 24.5 & 28.7 & & 65.7 & \textbf{71.4} & 82.0 \\
        \bottomrule
    \end{tabular}
    }
    \caption{s-BLEU, d-BLEU, and COMET scores of the Outside Context DocNMT models, tested with correct and random context. \textbf{Prev@2} and \textbf{Random@2} denote the previous two and random two-sentence context during the testing. The best scores are shown in bold.}
    \label{tab: main_results}
\end{table}

The COMET scores are also similar in the settings of both \textit{Prev@2} and \textit{Random@2}. On the News test set, \textit{MultiEnc-Mix@2} model achieves a score of 65.8 in \textit{Prev@2} setting. On the Ted test set, \textit{MultiEnc-Mix-Adapt@2} model obtains a score of 71.4 in \textit{Random@2} setting. Similarly, on the Europarl test set, \textit{MultiEnc-Mix@2} model achieves 82.1 in \textit{Prev@2} setting.
All the results indicate that the random context might not be random as the performance of the models is similar across both the context setting in terms of BLEU and COMET scores. To verify this, we conduct experiments on the ContraPro test set \citep{muller-etal-2018-large} to study the effects of random context and context encoder on pronoun translation accuracy.

\subsection{Results on the ContraPro test set}

ContraPro test set \citep{muller-etal-2018-large} is a test set for contrastive evaluation of models' performance on translating German pronouns \textit{es}, \textit{er} and \textit{sie}. Contrastive tests test the model's ability to discriminate between correct and incorrect outputs. In the ContraPro test set, the models' performance is measured in terms of the model's accuracy regarding reference pronoun, antecedent location, and antecedent distance. Similar to the training phase, we use the previous two sentences as context for the ContraPro test set, and Table \ref{tab: contrapro_results} shows the performance of the trained context-aware models. All the models' performance is similar to reference pronoun translation accuracy, with \textit{MultiEnc-Mix@2} and \textit{MultiEnc-Mix-Adapt@2} achieving a best overall score of 0.47. However, in terms of specific pronouns, \textit{MultiEnc-Random@2}, \textit{MultiEnc-Mix@2}, and \textit{MultiEnc-Mix-Adapt@2} achieved best scores of 0.87, 0.25, and 0.35 for pronouns \textit{es}, \textit{er} and \textit{sie} respectively. This shows that all the models can capture discourse information to translate pronouns, even the model trained with random context (\textit{MultiEnc-Random@2}).

\begin{table}[!ht]
    \centering
    \resizebox{1.0\linewidth}{!}{
    \begin{tabular}{lcccccccccccccc}
        \toprule
        \multirow{2}*{\textbf{Model}} & & & \multicolumn{3}{c}{reference pronoun} & & \multicolumn{2}{c}{antecedent location} & & \multicolumn{5}{c}{antecedent distance} \\
        \cmidrule{3-6}\cmidrule{8-9}\cmidrule{11-15}
         & & total & \textit{es} & \textit{er} & \textit{sie} & & intrasegmental & external & & 0 & 1 & 2 & 3 & $>$3 \\
        \midrule
        MultiEnc-Prev@2 & & 0.46 & 0.81 & 0.21 & 0.34 & & 0.71 & 0.39 & & 0.71 & 0.36 & 0.45 & 0.48 & 0.62 \\
        MultiEnc-Random@2 & & 0.46 & \textbf{0.87} & 0.18 & 0.33 & & 0.70 & 0.40 & & 0.70 & 0.36 & 0.46 & 0.49 & \textbf{0.68} \\
        MultiEnc-Mix@2 & & \textbf{0.47} & 0.85 & \textbf{0.25} & 0.31 & & 0.71 & \textbf{0.41} & & 0.71 & \textbf{0.38} & \textbf{0.47} & 0.48 & \textbf{0.68} \\
        MultiEnc-Mix-Adapt@2 & & \textbf{0.47} & 0.83 & 0.24 & \textbf{0.35} & & \textbf{0.73} & \textbf{0.41} & & \textbf{0.73} & \textbf{0.38} & 0.46 & \textbf{0.50} & 0.64 \\

        \bottomrule
    \end{tabular}
    }
    \caption{Accuracy on ContraPro test set for Outside Context DocNMT models regarding reference pronoun, antecedent location (within segment vs. outside segment), and antecedent distance of antecedent (in sentences). The best scores are shown in bold.}
    \label{tab: contrapro_results}
\end{table}

The results regarding antecedent location show that the \textit{MultiEnc-Mix-Adapt@2} model can perform well when the antecedent occurs in the current segment (intrasegmental). The \textit{MultiEnc-Prev@2} model can perform well when the antecedent is happening within the segment but poorly when the antecedent is outside (external). Similarly, \textit{MultiEnc-Random@2} model is able when the antecedent occurs outside the segment but poorly when the antecedent is within. However, both models \textit{viz.} \textit{MultiEnc-Mix@2} and \textit{MultiEnc-Mix-Adapt@2} perform well in both settings. The results show that mixing some random context is beneficial for the model to learn this discourse phenomenon effectively.

Similarly, the results regarding antecedent distance show that both the \textit{MultiEnc-Mix@2} and \textit{MultiEnc-Mix-Adapt@2} models achieve the best overall performance. Interestingly, the \textit{MultiEnc-Prev@2} model's performance is good when the antecedent distance is $<$3 but drops when the distance is $>$3, but \textit{MultiEnc-Random@2} model is achieving best score when the distance is $>$3 but slightly less in other settings than the \textit{MultiEnc-Prev@2} model. The results from \textit{antecedent distance} indicate that the model trained with random context can learn the long distant discourse properties better than the model trained with fixed context (previous two sentences). Based on this, we conclude that the random context might not be random as the model can learn discourse-level information well. These results also indicate that the context encoder might not generate noise as the discourse information outside the current source can only be learned if the context encoder can encode the context well. We further investigate this by analyzing the source sentence embeddings.

\subsection{t-SNE Visualization of Source and Target Embeddings}

\begin{figure}[!ht]
    \centering
    \subfigure[]{\includegraphics[scale=0.40]{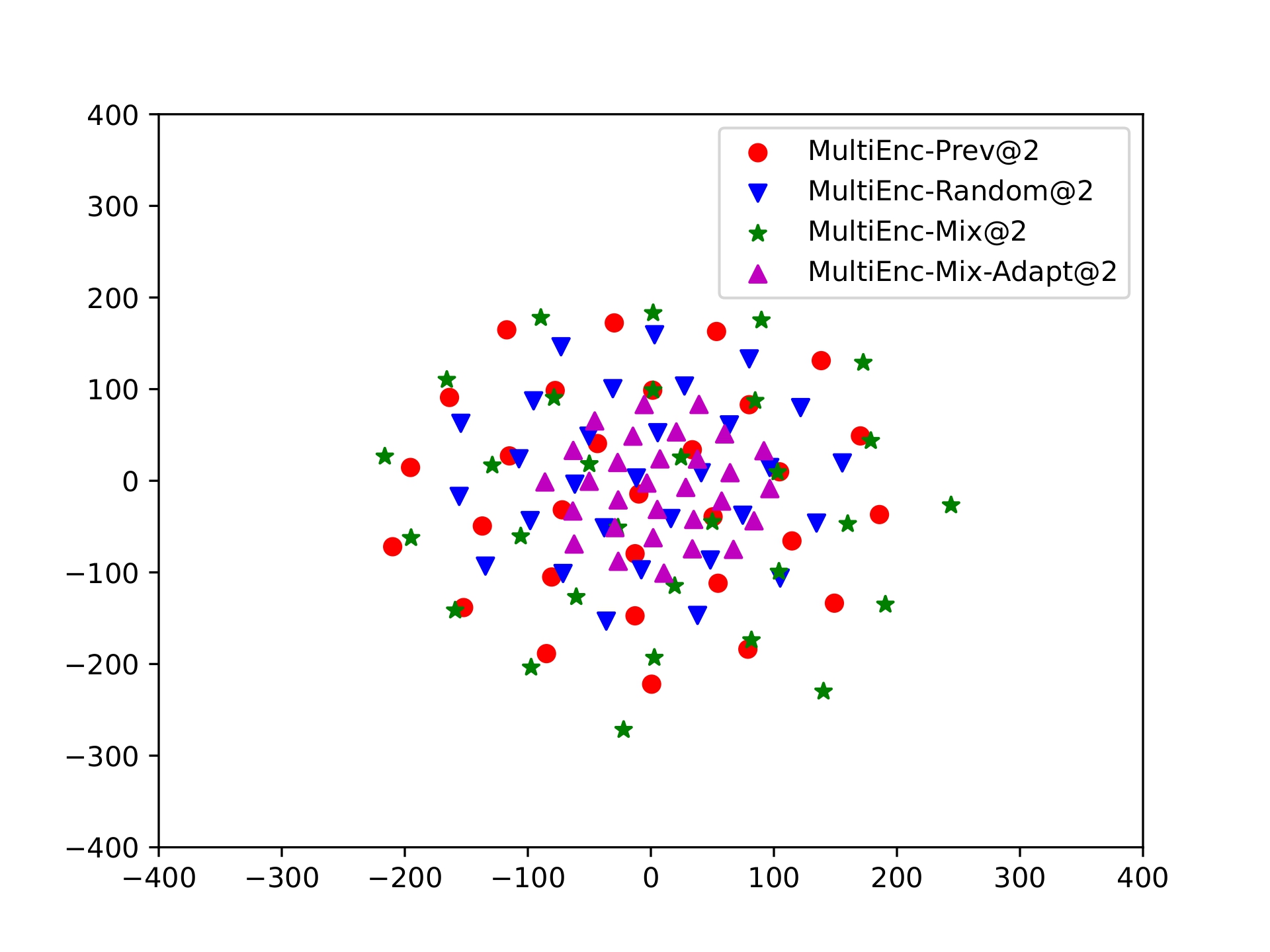}}
    \subfigure[]{\includegraphics[scale=0.40]{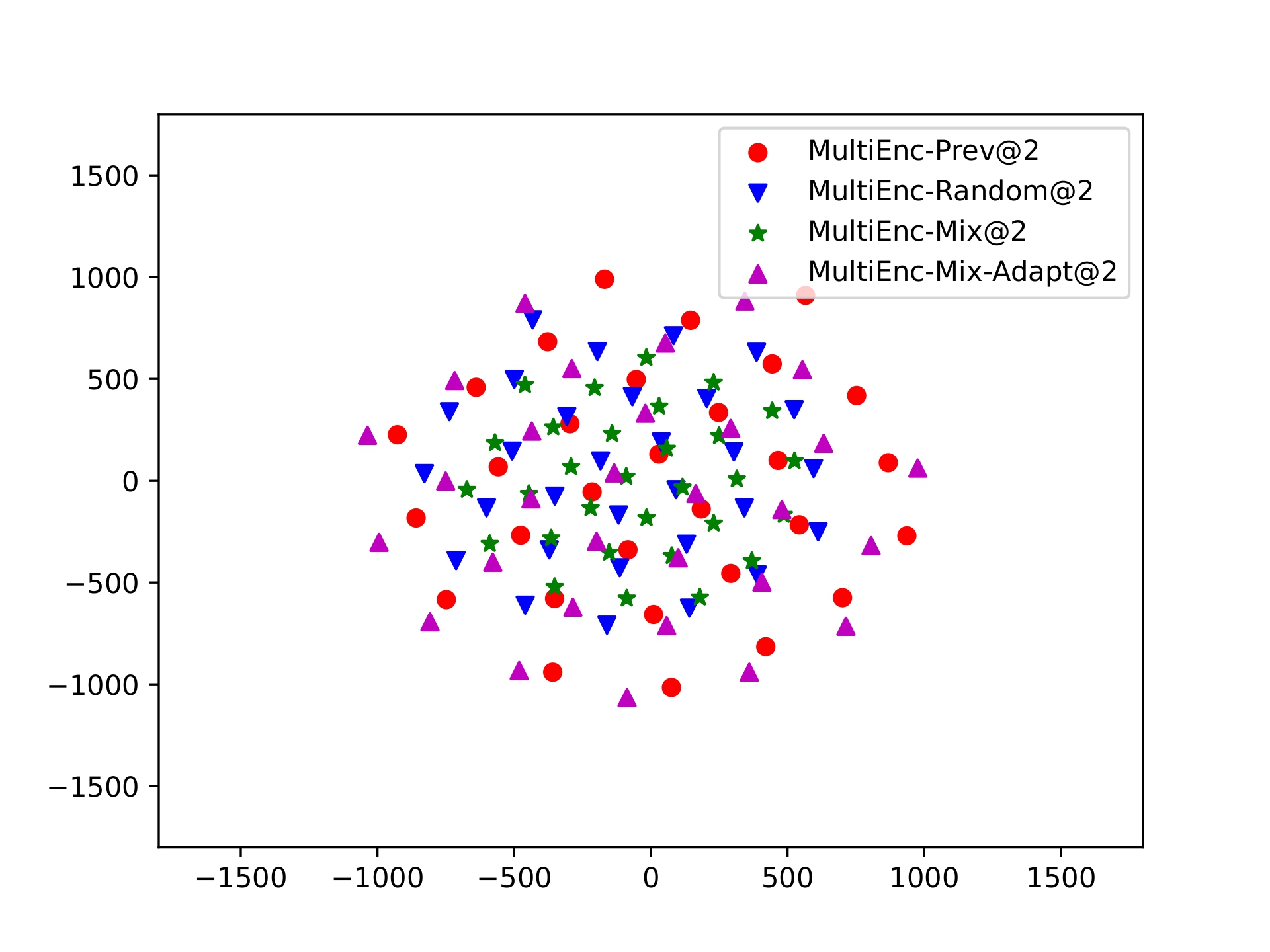}}
    \subfigure[]{\includegraphics[scale=0.40]{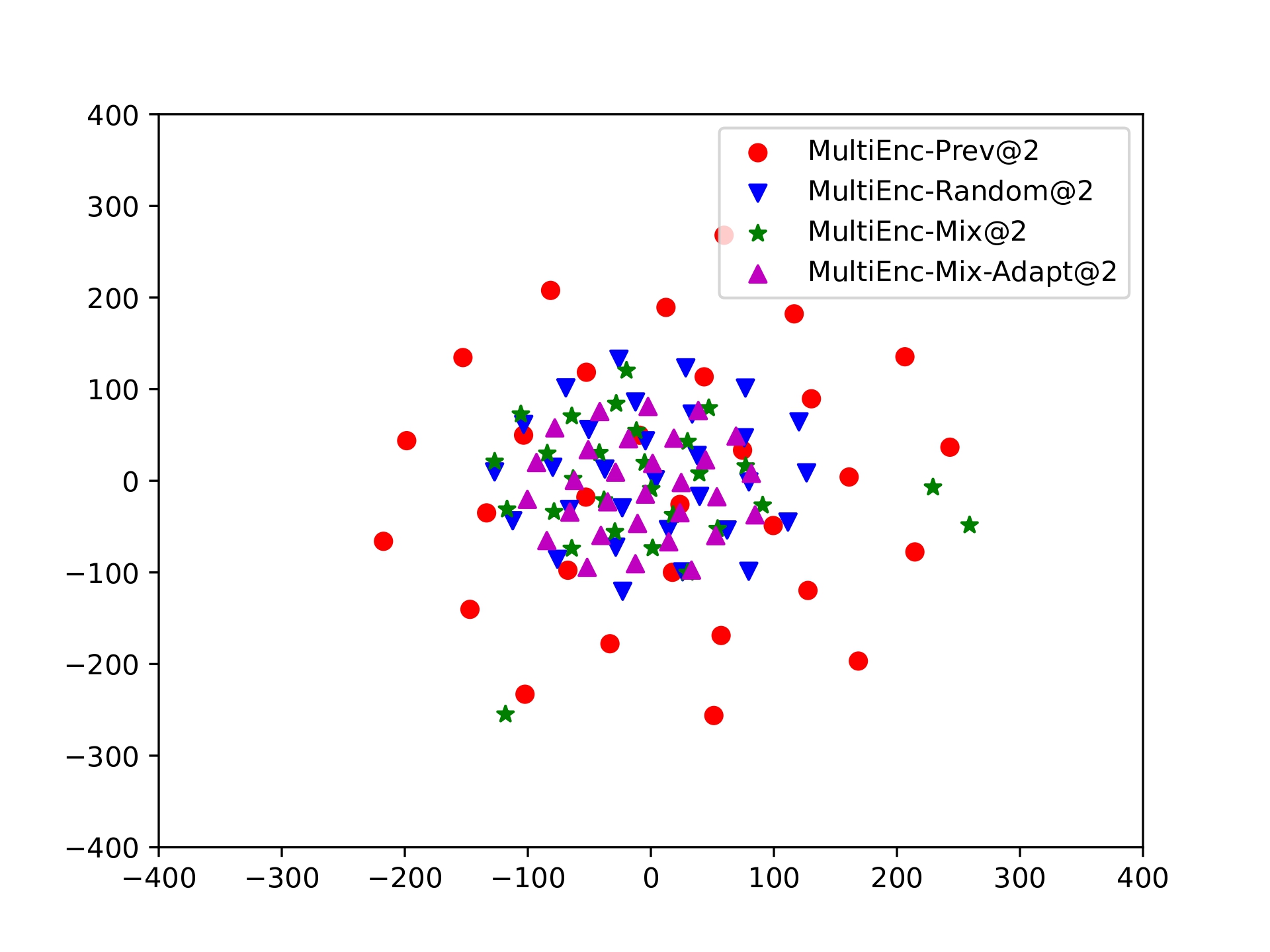}}
    \subfigure[]{\includegraphics[scale=0.40]{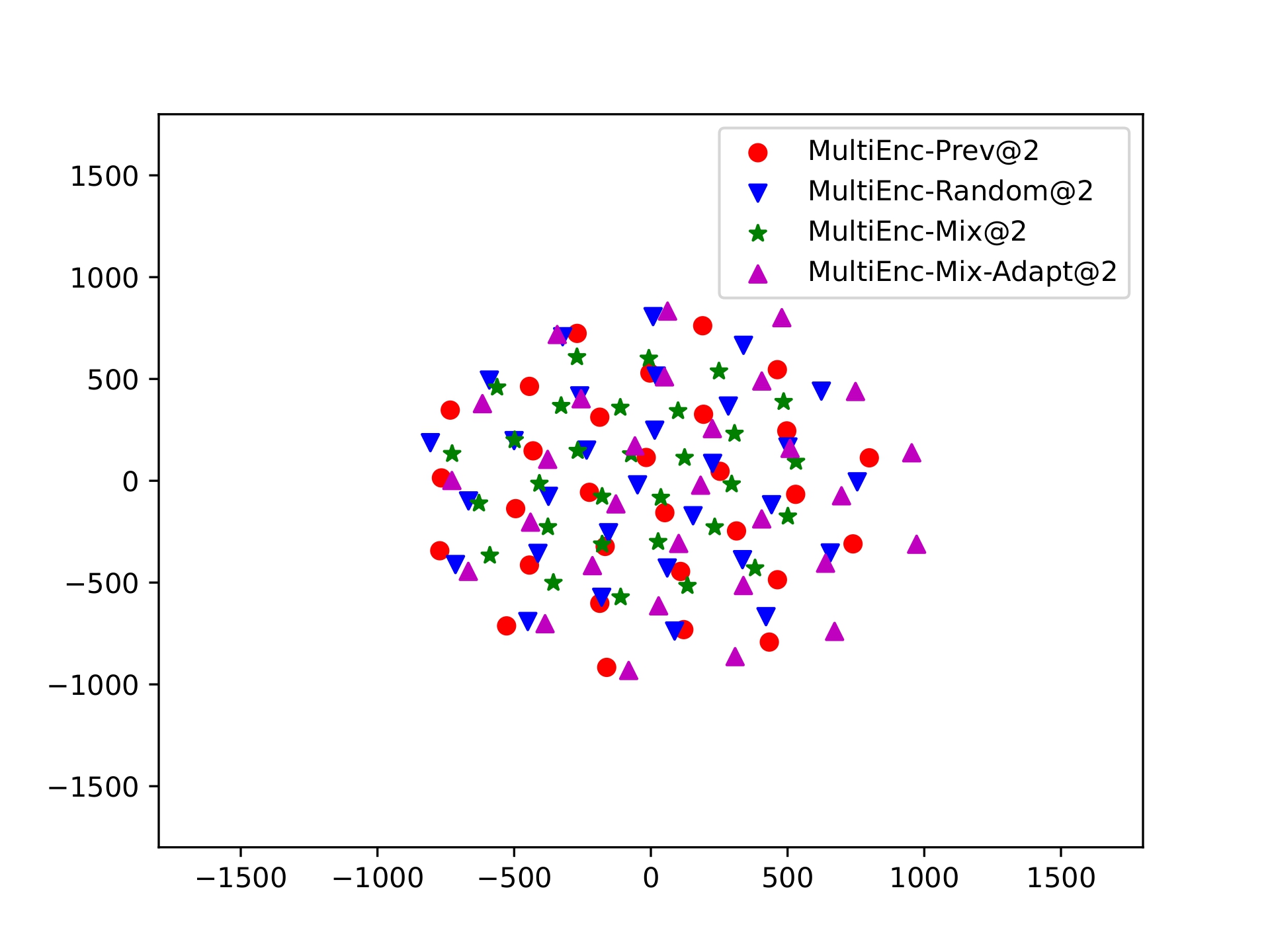}}
    \caption{t-SNE visualization of the source and target representations of the Outside Context Multi-Encoder DocNMT models with different context settings. (a) and (b) show when the previous two sentences are used as context, (c) and (d) show When two random sentences are used as context. (a) and (c) show the source representations, (b) and (d) shows the target representations. Each point represents a sentence.}
    \label{fig: sentence_embeddings}
\end{figure}

Since the relation between the sentences in a given document is learned through context, the context encoder should be trained sufficiently to capture this aspect. To study this, we take a document and visualize the source sentence representations obtained after performing attention over context and source encoder outputs and combining the resulting output with source encoder output via element-wise addition (cf. Fig \ref{fig: outside-attn}). Figure \ref{fig: sentence_embeddings} shows t-SNE visualization \citep{van2008visualizing}\footnote{\url{https://scikit-learn.org/stable/modules/generated/sklearn.manifold.TSNE.html}} of the source representations. The document is taken from the train set of \textit{News-commentary v14} corpus and contains 30 sentences. We obtain the source representations in two settings similar to the testing phase \textit{viz.} previous two sentences and randomly sampled two sentences as context.

Figure \ref{fig: sentence_embeddings}(a) shows the representation when the context consists of the previous two sentences. Interestingly the representations from \textit{MultiEnc-Prev@2} and \textit{MultiEnc-Mix@2} models spread out more than the other two models. The reason might be that the model requires more context to encode the sentences effectively for these two models. The \textit{MultiEnc-Random@2} model can encode the sentences better than the \textit{MultiEnc-Prev@2} model. This might be why the models trained with random context can perform well, as the model can learn sufficient information even from the random context. This shows that the random context might not be random, and it is helping the model to encode sentences well enough to capture the discourse-level information. The \textit{MultiEnc-Mix-Adapt@2} model can learn the source representation better than other models as the sentences are projected in a smaller zone. This shows that the \textit{MultiEnc-Mix-Adapt@2} model can perform well even when the context is limited.

Similarly, Figure \ref{fig: sentence_embeddings}(c) shows the representation when the context consists of two random sentences. The representation of \textit{MultiEnc-Prev@2} model is adversely affected by the random context, but all other models can learn the representations of sentences well. Interestingly, the spread of \textit{MultiEnc-Mix@2} model is smaller than the spread when the context consists of the previous two sentences. This indicates that the model can learn better representations from random context even though the model is trained by mixing 50\% of previous sentence context and 50\% of random context. The same is true for the \textit{MultiEnc-Random@2} model also. The representations of the \textit{MultiEnc-Mix-Adapt@2} model are the same as the model trained with the previous sentence context and show that this model is consistent even when the type of context changes.

We also analyze how different types of context affect the decoder. Figure \ref{fig: sentence_embeddings}(b) and Figure \ref{fig: sentence_embeddings}(d) shows the target representations when the context consists of the previous two and random two sentences, respectively. We observe that the source context is insufficient to project the target embeddings closer even though the sentences are from the same document. Interestingly, in the random context setting (cf. Figure \ref{fig: sentence_embeddings}(d)), the representations are projected closer than the correct context setting (cf. Figure \ref{fig: sentence_embeddings}(b)), indicating that the decoder is mainly unaffected by the choice of the context. This might be due to the context being chosen from the source side and leading to the sub-optimal encoding of target sentences. We hypothesize that the model can learn better target representation if trained with target-side context.

Based on the analysis, the model can learn discourse-level properties even when the context is random, indicating that the random context might not adversely affect the context-aware model. This also shows that the context encoder might not be generating noise; instead, it can generate sufficient information to capture discourse-level properties based on the type of context the model is trained with. The t-SNE visualization shows that the source representations are affected by the choice of context, and the target representations are mainly unaffected. This suggests that metrics such as BLEU might not be enough to measure the discourse-level information the system can learn and requires unique discourse-level test sets to evaluate.

\begin{table}[!ht]
    \centering
    \small
    \begin{tabular}{lccc}
        \toprule
        \textbf{Model} & \textbf{News} & \textbf{TED} & \textbf{Europarl} \\
        \midrule
        MultiEnc-Prev@2 & 16.9 & 16.1 & 22.8 \\
        MultiEnc-Random@2 & 14.1 & 15.7 & 20.1\\
        MultiEnc-Mix@2 & 20.5 & 19.0 & 24.4 \\
        MultiEnc-Mix-Adapt@2 & \textbf{23.0} & \textbf{20.5} & \textbf{26.5} \\
        \bottomrule
    \end{tabular}
    \caption{s-BLEU scores of the Outside Context DocNMT models, tested with the same source sentences as the context. The best scores are shown in bold.}
    \label{tab: sst_results}
\end{table}

\subsection{Results of Multi-Encoder models with identical Source and Context}

We further study whether the context encoder generates noise by feeding the same source sentence as the context. If the context encoder is generating noise, then the models' performance should be similar, as the inputs are identical to every model. Table~\ref{tab: sst_results} shows the s-BLEU scores of the models tested in this setting. Interestingly \textit{MultiEnc-Random@2} models' performance is lowest than the other models. This indicates that the model becomes sensitive to the choice of the context when the context is random. Similarly, \textit{MultiEnc-Mix@2} and \textit{MultiEnc-Mix-Adapt@2} models' performance is better than the other models. Results indicate that mixing the random context with the correct context makes the model robust and results in better-quality translations.

\section{Conclusion and Future Work}

In this work, we conducted experiments on multi-encoder-based DocNMT systems to study how different types of contexts affect context-aware pronoun translation. Specifically, we consider three different types of context settings \textit{viz,} previous two sentences, random two sentences, and a mix of both these settings. We use the ContraPro test set as the context-aware test set to analyze the pronoun translation accuracy. Our analysis shows that the multi-encoder models can perform well on pronoun translation even when the context is random. We further conduct experiments to study whether the context encoder is generating noise or not by projecting the sentence representations from a single document using t-SNE. The analysis shows that the context encoder can encode the context sufficiently enough to capture the relation between the sentences, as these sentences are connected only via context. Based on the analysis, we conclude that the random context might not adversely affect the performance of multi-encoder-based DocNMT models. Choosing context is essential for effectively capturing any discourse phenomenon. The context encoder might not be generating noise. Instead, the encoding from the context encoder is dependent on the choice of context. As we observed that mixing selected context (previous two sentences in this case) and random context is performing better than the other settings, we plan to explore effective context encoding through contrastive learning \citep{hwang-etal-2021-contrastive} and dynamic context generation based on the source and target pairs which can help during the inference in round-trip-translation \citep{tu2017neural} method.

\section*{Acknowledgements}
We gratefully acknowledge the support from ``NLTM: VIDYAAPATI'' project, sponsored by Electronics and IT, Ministry of Electronics and Information Technology (MeitY), Government of India. We also thank the anonymous reviewers for their insightful comments.

\small

\bibliographystyle{apalike}
\bibliography{mtsummit2023, anthology}


\end{document}